\documentclass[runningheads]{llncs}
\usepackage{graphicx}
\usepackage{subcaption}
\usepackage{amsmath}
\usepackage{amssymb}
\usepackage{comment}
\usepackage{xcolor}
\definecolor{matlabBlue}{RGB}{0,114,189}
\definecolor{matlabRed}{RGB}{217,83,24}
\definecolor{matlabYellow}{RGB}{236,176,32}

\usepackage[hidelinks=true,breaklinks=true]{hyperref}
\usepackage[capitalise]{cleveref}
\crefname{equation}{}{}
\Crefname{equation}{}{}

\begin{document}
\title{Robot Path and Trajectory Planning Considering a Spatially Fixed TCP\thanks{This work has been supported by the ”LCM - K2 Center for Symbiotic Mechatronics” within the framework of the Austrian COMET-K2 program.}}

\author{Bernhard Rameder\inst{1}\orcidID{0000-0002-6792-6129} \and
Hubert Gattringer\inst{1}\orcidID{0000-0002-8846-9051} \and
Andreas Müller\inst{1}\orcidID{0000-0001-5033-340X} \and
Ronald Naderer\inst{2}}
\authorrunning{B. Rameder et al.}
%
\institute{Institute of Robotics, Johannes Kepler University Linz\\Altenberger Straße 69, 4040 Linz, Austria
	\email{\{bernhard.rameder,hubert.gattringer,a.mueller\}@jku.at}\\
	\and
	FerRobotics Compliant Robot Technology GmbH\\Altenberger Straße 66c,  4040 Linz, Austria \\
	\email{ronald.naderer@ferrobotics.at}
} 
\maketitle              
\begin{abstract}
This paper presents a method for planning a trajectory in workspace coordinates using a spatially fixed tool center point (TCP), while taking into account the processing path on a part. This approach is beneficial if it is easier to move the part rather than moving the tool. Whether a mathematical description that defines the shape to be processed or single points from a design program are used, the robot path is finally represented using B-splines. The use of splines enables the path to be continuous with a desired degree, which finally leads to a smooth robot trajectory. While calculating the robot trajectory through prescribed orientation, additionally a given velocity at the TCP has to be considered. The procedure was validated on a real system using an industrial robot moving an arbitrary defined part.

\end{abstract}
\section{Introduction}
\begin{figure}[htb]
	\centering
	\begin{subfigure}{0.49\textwidth}
		\centering
		\includegraphics[height=4.5cm]{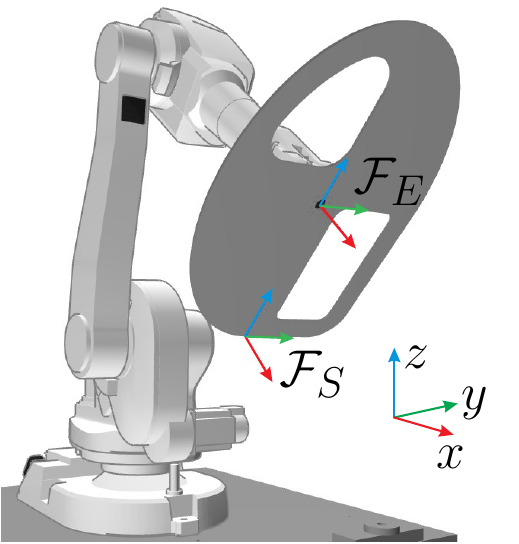}	
	\end{subfigure}
	\hfill		
	\begin{subfigure}{0.49\textwidth}
		\centering
		\includegraphics[height=4.5cm]{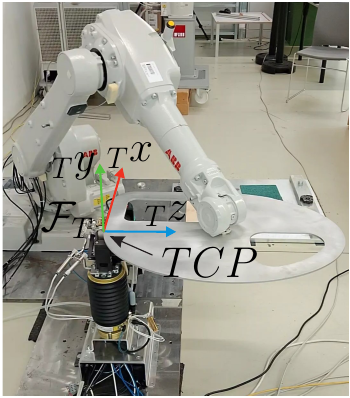}	
	\end{subfigure}
	\caption{Robot with exemplary part. Part borders represent processing path.}
	\label{fig:Part}
\end{figure}

Due to the increasing automation of industry, many processes today are supported by industrial robots to make them either faster or more efficient. In many cases the robot guides the tool along a defined path to process a certain part, see e.g. \cite{amersdorfer2021equidistant}. However, this may not always be the best solution, especially if the used tool is relatively bulky compared to the part to be worked on. This paper therefore describes an approach for calculating a suitable robot trajectory that guides the part, instead of the tool, along the required, spatially fixed tool, depending on its position and orientation, taking into account the processing path on the part. A similar approach was shown in \cite{hartl2020surface}, where the authors described the application of tapes on complex 3D objects. In general, the handling of the part can be beneficial for tasks where the part is either rather small and complex compared to the tool or where the tool needs many additional wiring or consumables to work sufficiently. This extra equipment may constrain the motion of the robot and therefore the overall workspace, whereas an appropriate part handling may not restrict the motion at all. Processes that can benefit from the trajectory calculation approach presented in this paper are for instance sealant application tasks and the automated tape laying or welding process, where the tool has to be provided with operating materials like the sealant, the tape, welding gas or the welding wire. Additionally, these tasks often requires connections to compressed air, hydraulics or to electricity to be able to control the functions of the tool. Many approaches deal with path planning using CAD data, but do not consider a spatially fixed tool, as it is the case in \cite{HepingChen2008,chen2008automated}, where different manufacturing paths are automatically planned for a robot end effector fixed TCP. To overcome the named issues, this paper introduces a way to get from a geometrical representation of a processing path on a part to a robot trajectory, which furthermore considers a given velocity at the TCP. In the following, the geometric description of the part with B-splines, the calculation of the needed robot path as well as the trajectory calculation will be handled. The method was finally tested on an industrial robot using an arbitrary defined part. The part and the experimental setup are shown in \cref{fig:Part}, where $\mathcal{F}_E$ represents the robot end effector frame, $\mathcal{F}_S$ the frame of the starting point of the processing path and $\mathcal{F}_T$ the tool frame located at the TCP.

\section{Geometric Description of Processing Path on Part}
As already mentioned, the desired shape of the processing path from the part can either be described using a mathematical description or by using certain interpolation/approximation points from a design program like CAD. Nevertheless, both representations benefit from a depiction via B-splines of at least fifth order, if the jerk of the trajectory should be a smooth function, as the authors of \cite{gasparetto2007new} did for a smooth trajectory planning for robot manipulators. The first step is therefore to reformulate the geometry of the path on the part into B-splines using given points $\mathbf{r}_{j}(\sigma_j)$ at particular parameter values $\sigma_j$ with $j = 0, \hdots, N-1$. These points $\mathbf{r}_j$ are either given directly from the obtained points from CAD or by sufficient sampling of the mathematical description. Using the control points $\mathbf{d}_{i}$ combined with the piecewise polynomial function $N_{i,p}(u)$ \cite{piegl1997nurbs} of a B-spline with degree $p$ and a knot vector $(u_0, \hdots, u_{n-1})$ with length $n$, the processing path can be described as B-spline curve
\begin{equation}
	{_P}\mathbf{r}_{PF}(u) = \sum^{m-1}_{i=0}\mathbf{d}_{i} N_{i,p}(u), u \in [a,b]
\end{equation}
in part frame $\mathcal{F}_P$, where $m = n-p-1$ is the number of B-spline basis functions. The allowed interval of the knot value $u$ is defined by $a = u_0 = \hdots = u_p$ and $b = u_{n-p-1} = \hdots = u_{n-1}$. ${_P}\mathbf{r}_{PF}$ denotes a vector from point $P$ to point $F$, represented in frame $\mathcal{F}_{P}$, which is indicated by the left subscript. The path parameter $a \leq \sigma_k \leq b$ in this case is chosen as
\begin{equation}
	\sigma_{k} = a + \frac{b -a}{L} \sum^{k - 1}_{j = 0} \| \mathbf{r}_{j+1} - \mathbf{r}_{j} \|
	\label{eq:sigma_k}
\end{equation}
with $k = 1, \hdots, N-1$ and $L = \sum^{N - 2}_{j = 0} \| \mathbf{r}_{j+1} - \mathbf{r}_{j} \|$ being the length of the data polygon. If the number of given points $\mathbf{r}_j(\sigma_j)$ with $j = 0, \hdots, N-1$ is higher than the number of basis functions $N \geq m$ the problem can be solved with a B-spline approximation using the weighted least squares method
\begin{align}
	&\underset{\mathbf{d}_{i}}{\min} \, \sum^{N-1}_{j=0} w_j \Biggl\| \sum^{m-1}_{i=0} \mathbf{d}_i N_{i,p}(\sigma_{j}) - \mathbf{r}_{j} \Biggr\|^2, 
\end{align}
where the sum of the quadratic distance of the points $\mathbf{r}_{j}$ to the spline curve is minimized \cite{YuChengHwang2008,piegl1997nurbs,Smyth2002} by finding an optimal set of control points $\mathbf{d}_{i}$. Therefore, the \textit{Matlab} function \textit{spap2} is used.
For the following descriptions the exemplary test part, depicted in \cref{fig:Part}, is considered.

\section{Robot Path Planning}
Usually, an appropriate robot trajectory can easily be calculated using the desired processing path ${_P}\mathbf{r}_{PF}$ on a part, described in frame $\mathcal{F}_P$, where the part is defined. However, things are different, if the TCP is spatially fixed. The described path on the part has to be translated into a path for the robot, which considers the position and orientation of the spatially fixed tool and therefore leads to a totally different path for the robot compared to the one on the part. An important role for calculating the needed path plays the process orientation frames, which are defined and handled in more detail in the following.

\subsection{Calculation of the Process Orientation Frames}
\begin{figure}[htb]
	\centering
	\includegraphics[height=5cm,trim={0 1.5cm 0 0.5cm},clip]{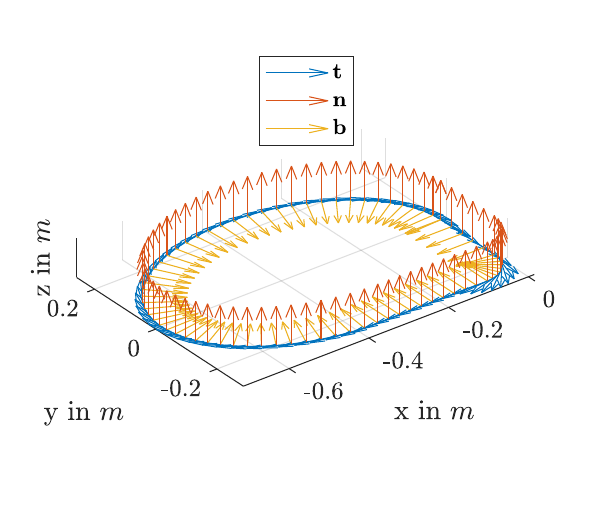}	
	\caption{Process orientation frames $\mathcal{F}_F$ depicted along processing path ${_P}\mathbf{r}_{PF}$}
	\label{fig:OrientationFrames}
\end{figure}

The Frenet frame \cite{siciliano2008Robotics} is of central importance for calculating the orientation of a tool on a robot path in general. Furthermore, in the case of a spatially fixed tool, this frame is necessary to determine the robot path itself as well as the corresponding orientation, starting from a predefined processing path on a part. Normally, the Frenet frame is composed out of the tangential, normal and binormal vector, defined by the geometry of the path. However, in some cases it can be necessary to adapt this frame to fulfill special characteristics for a certain path or tool. For instance, if a roller as a tool is used, it may be reasonable to define the tangential vector
\begin{equation}
	{_P}\mathbf{t}(\sigma) = \frac{1}{\Big\|\frac{\partial{{_P}\mathbf{r}}_{PF}(\sigma)}{\partial{\sigma}}\Big\|} \frac{\partial{{_P}\mathbf{r}}_{PF}(\sigma)}{\partial{\sigma}}
	\label{eq:tangVec}
\end{equation}
of the path along the free degree of freedom ${_T}x$ (\cref{fig:Part}) of the roller in the desired contact point. The normal vector ${_T}y$ of the roller surface in the contact point should be aligned with the known normal vector ${_P}\mathbf{n}_{S}(\sigma)$ pointing into the surface where the roller gets pressed on and hence should be aligned with the normal vector of the path ${_P}\mathbf{n}(\sigma) = {_P}\mathbf{n}_{S}(\sigma)$. The binormal vector ${_{P}}\mathbf{b}$ of the path is then automatically defined using the cross product ${_P}\mathbf{b}(\sigma) = {_P}\mathbf{t}(\sigma) \times {_P}\mathbf{n}(\sigma)$ of tangential ${_P}\mathbf{t}(\sigma)$ and normal vector ${_P}\mathbf{n}(\sigma)$. In contrast to the Frenet frame, this adaption is named process orientation frame $\mathcal{F}_F$ in the following and defines the alignment of the tool at the single points along the processing path. The corresponding rotation matrix is $\mathbf{R}_{PF} = [{_{P}}\mathbf{t}, {_{P}}\mathbf{n}, {_{P}}\mathbf{b}]$ and represents a transformation from the current process orientation frame $\mathcal{F}_F$ into frame $\mathcal{F}_P$, where the part geometry is defined. This matrix is later used for calculating the desired robot path ${_I}\mathbf{r}_{0E}$ using the given processing path ${_P}\mathbf{r}_{PF}$ on the part. According to the used part in \cref{fig:Part} the calculated orientation frames $\mathcal{F}_F$ along the processing path ${_P}\mathbf{r}_{PF}$ are presented in \cref{fig:OrientationFrames}.

\subsection{Calculation of the Robot Path Using the TCP Location and Processing Path Geometry on the Part}
\begin{figure}[htb]
	\centering
	\begin{subfigure}{0.49\textwidth}
			\centering
			\includegraphics[height=4.9cm]{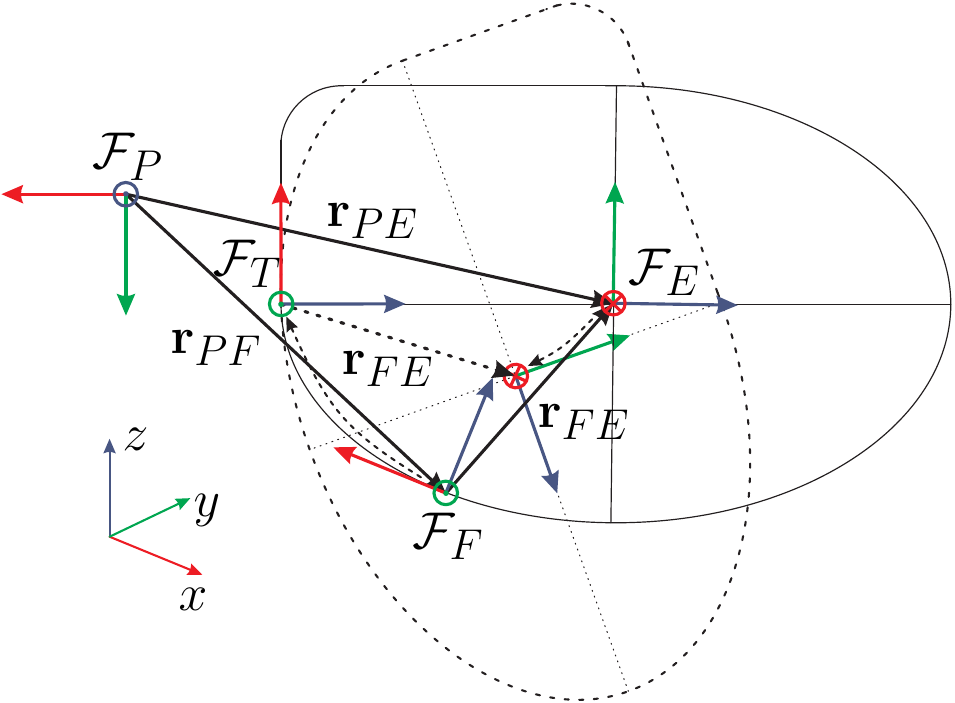}	
			\subcaption{End effector motion relative to $\mathcal{F}_F$}
			\label{fig:RobotPathCalculationFrameP}
		\end{subfigure}
	\hfill		
	\begin{subfigure}{0.49\textwidth}
			\centering
			\includegraphics[height=6cm]{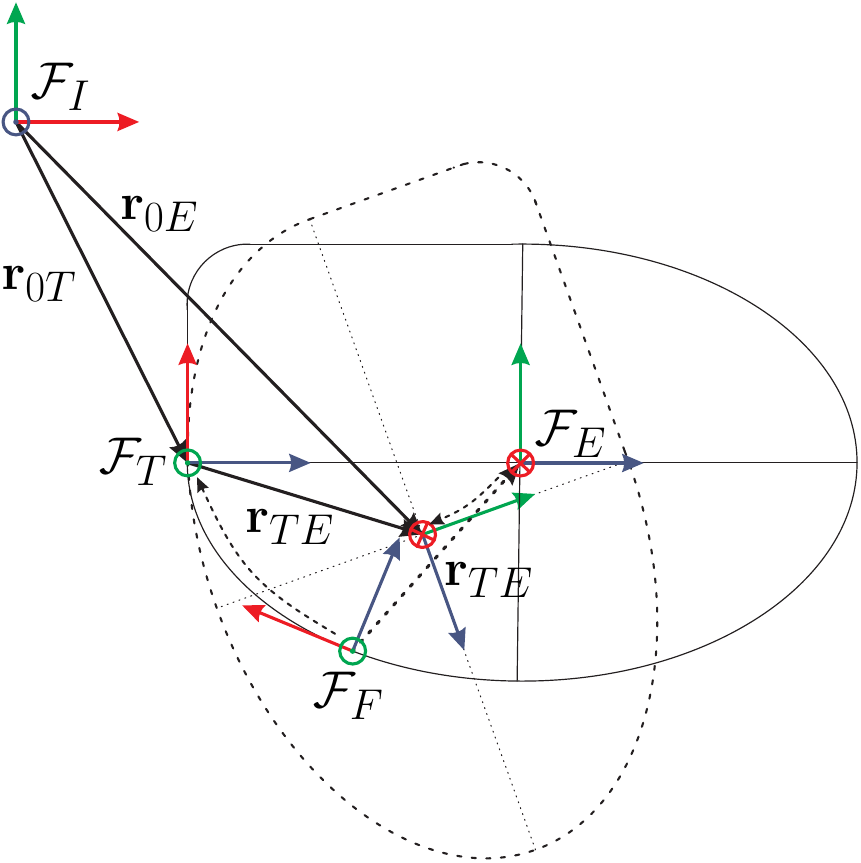}	
			\subcaption{End effector motion relative to $\mathcal{F}_I$}
			\label{fig:RobotPathCalculationFrameI}
		\end{subfigure}
	\caption{Graphical representation of the robot end effector path calculation}
	\label{fig:RobotPathCalculation}
\end{figure}

\begin{figure}[htb]
 	\centering
 	\begin{subfigure}{0.49\textwidth}
 		\centering
 		\includegraphics[height=4.5cm,trim={0 0 0 0cm}]{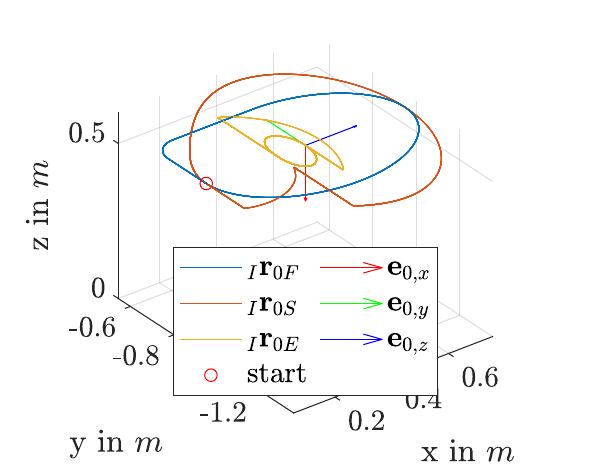}	
 		\subcaption{Processing path and robot path}
 		\label{fig:ProcessingAndRobotPathPosition}
 	\end{subfigure}
 	\hfill		
 	\begin{subfigure}{0.49\textwidth}
 		\centering
 		\includegraphics[height=5cm,trim={0 1.5cm 0 0cm}]{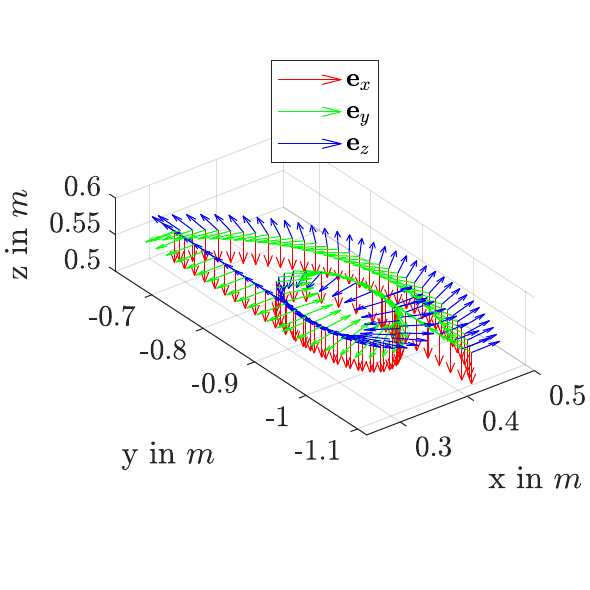}	
 		\subcaption{Robot path with orientation}
 		\label{fig:RobotPathOrientation}
 	\end{subfigure}
 	\caption{Calculated robot path ${_I}\mathbf{r}_{0E}$ and corresponding orientation $\mathbf{R}_{IE}$}
 	\label{fig:RobotPath}
 \end{figure}

Once the position ${_P}\mathbf{r}_{PF}$ and orientation $\mathbf{R}_{PF}$ of the single points on the processing path are known, the corresponding robot path ${_I}\mathbf{r}_{0E}$ and $\mathbf{R}_{IE}$ can be calculated. Assuming ${_P}\mathbf{r}_{PF}$ and $\mathbf{R}_{PF}$ represent the position and orientation of the tool in relation to frame $\mathcal{F}_P$, frame $\mathcal{F}_F$ can be replaced by frame $\mathcal{F}_T$, which leads to ${_P}\mathbf{r}_{PT} \stackrel{!}{=} {_P}\mathbf{r}_{PF}$ and $\mathbf{R}_{PT} \stackrel{!}{=} \mathbf{R}_{PF}$. If this relation is seen with respect to a spatially fixed frame $\mathcal{F}_T$, the trajectory along the processing path can be seen as rotation of the part with respect to $\mathcal{F}_T$, which is indicated by the dashed line in \cref{fig:RobotPathCalculation}. In order to obtain a relation to the robot end effector with frame $\mathcal{F}_E$, the constant position ${_P}\mathbf{r}_{PE}$ and mounting direction $\mathbf{R}_{PE}$ of the end effector on the part in frame $\mathcal{F}_P$ is defined, which can be seen in \cref{fig:RobotPathCalculationFrameP}.
Now it is possible to define a vector between the tool frame $\mathcal{F}_T$ and the end effector frame $\mathcal{F}_E$ using
\begin{equation}
	{_P}\mathbf{r}_{TE} ={_P}\mathbf{r}_{PE} - {_P}\mathbf{r}_{PT}.
\end{equation}
The corresponding relative rotation is obtained using the equation
\begin{equation}
	\mathbf{R}_{TE} = \mathbf{R}_{TP} \mathbf{R}_{PE}.
\end{equation}
If ${_P}\mathbf{r}_{TE}$ is evaluated in frame $\mathcal{F}_T$ using ${_T}\mathbf{r}_{TE} = \mathbf{R}_{TP} {_P}\mathbf{r}_{TE}$, which is equivalent to the end effector motion relative to $\mathcal{F}_T$, it is easy to compute this vector in the inertial frame $\mathcal{F}_I$ of the robot without knowing the position and orientation of frame $\mathcal{F}_P$ related to frame $\mathcal{F}_I$. Therefore, the known and spatially fixed position ${_I}\mathbf{r}_{0T}$ and orientation $\mathbf{R}_{IT}$ of the tool with respect to the inertial system $\mathcal{F}_I$ of the robot is used, which leads to the needed robot path represented in frame $\mathcal{F}_I$ including position
\begin{equation}
 	{_I}\mathbf{r}_{0E} = {_I}\mathbf{r}_{0T} + \mathbf{R}_{IT} {_T}\mathbf{r}_{TE}
\end{equation}
and orientation
\begin{equation}
 	\mathbf{R}_{IE} = \mathbf{R}_{IT} \mathbf{R}_{TE}.
\end{equation}
According position vectors and orientation frames are shown in \cref{fig:RobotPathCalculationFrameI} for a certain point in time.
The result for the position of the robot end effector path ${_I}\mathbf{r}_{0E}$ and the original processing path ${_I}\mathbf{r}_{0F}$ with respect to frame $I$ is shown in \cref{fig:ProcessingAndRobotPathPosition}. Additionally, the temporal development of the starting point from the processing path, located in the center of $\mathcal{F}_S$, is represented as path ${_I}\mathbf{r}_{0S}$ in this figure. The unit vectors $(\mathbf{e}_{0,x},\mathbf{e}_{0,y},\mathbf{e}_{0,z})$ in \cref{fig:ProcessingAndRobotPathPosition} represent the orientation of frame $\mathcal{F}_E$ at the starting point of the path. The robot path with corresponding orientations, depicted as frames $\mathcal{F}_E$ composed of the unit vectors $(\mathbf{e}_{x},\mathbf{e}_{y},\mathbf{e}_{z})$ of $\mathbf{R}_{IE}$, is shown in \cref{fig:RobotPathOrientation}.

\section{Robot Trajectory Calculation}
\begin{figure}[htb]
	\centering
	\begin{subfigure}{0.49\textwidth}
		\centering
		\includegraphics[height=6.5cm,trim={0 0.75cm 0 0.75cm}]{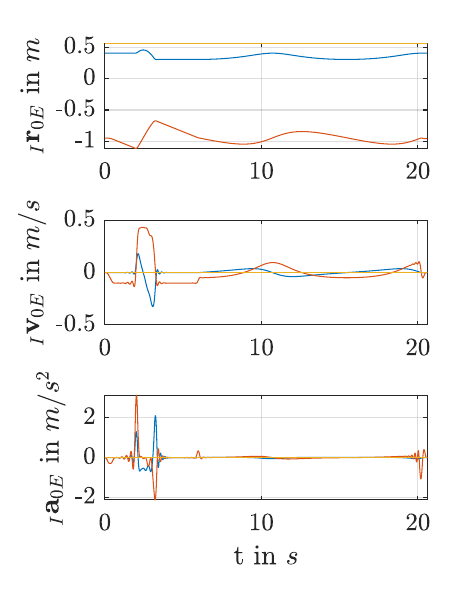}
		\label{fig:robTrajPos}
	\end{subfigure}
	\hfill		
	\begin{subfigure}{0.49\textwidth}
		\centering
		\includegraphics[height=6.5cm,trim={0 0.75cm 0 0.75cm}]{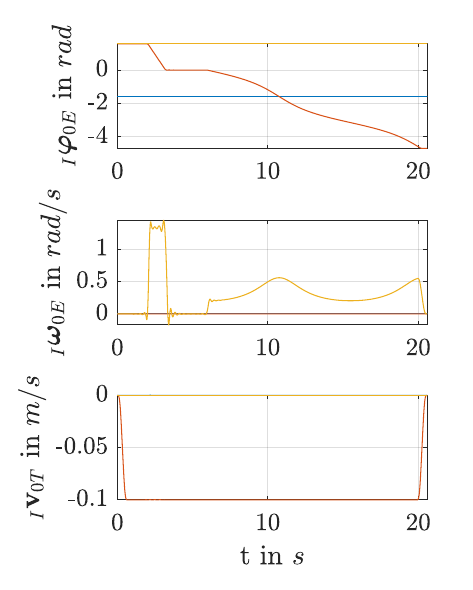}
		\label{fig:robTrajOrient}
	\end{subfigure}
	\caption{Calculated robot trajectory considering a spatially fixed TCP \\
	\centering Legend: \textcolor{matlabBlue}{------} x or $\alpha$, \textcolor{matlabRed}{------} y or $\beta$, \textcolor{matlabYellow}{------} z or $\gamma$}
	\label{fig:robotTrajSin2}
\end{figure}

To be able to process the calculated path ${_I}\mathbf{r}_{0E}$ on a robotic system, it is necessary to add an appropriate time information to get a trajectory by defining the path parameter $\sigma(t)$. One option is to create piecewise $sin^2$ trajectories for $\dddot{\sigma}(t)$, taking the maximum velocity, acceleration and jerk into consideration, based on applications used in \cite{biagiotti2008trajectory}. To ensure the defined maximum values for velocity, acceleration and jerk of $\sigma$ to be directly connected with the ones of the final trajectory, the path parameter has to be arc length parameterized along the path, which is approximated using \cref{eq:sigma_k}. This is useful, if, for instance, the processing speed for a certain task is important, like it is for welding or taping. Depending on whether the velocity at the TCP or at the end effector should be maintained, $\sigma$ has to be either arc length parameterized along the original processing path ${_P}\mathbf{r}_{PF}$ or along the new robot path ${_I}\mathbf{r}_{0E}$, respectively. The corresponding poses $\mathbf{z}^T_{E} = [{_I}\mathbf{r}^T_{0E} \ {_I}{\boldsymbol{\varphi}}^T_{0E}]$, velocities $\dot{\mathbf{z}}^T_{E} = [{_I}{\mathbf{v}}^T_{0E} \ {_I}{\dot{\boldsymbol{\varphi}}}^T_{0E}]$ and accelerations $\ddot{\mathbf{z}}^T_{E} = [{_I}{\mathbf{a}}^T_{0E} \ {_I}{\ddot{\boldsymbol{\varphi}}}^T_{0E}]$ in workspace coordinates $\mathcal{F}_I$ are evaluated with the equations
\begin{equation}
	\mathbf{z}_E = \mathbf{z}_E(\sigma), \ 
	\dot{\mathbf{z}}_E = {\frac{\partial \mathbf{z}_E}{\partial \sigma}} \dot{\sigma} \ \text{and } 
	\ddot{\mathbf{z}}_E = {\frac{\partial^2 \mathbf{z}_E}{\partial \sigma^2}} \dot{\sigma}^2 + {\frac{\partial \mathbf{z}_E}{\partial \sigma}}\ddot{\sigma},
\end{equation}
like it is done in \cite{Pham2014}. Thereby, the orientation is described in Tait-Bryan angles ${_I}{\boldsymbol{\varphi}}^T_{0E}$. These angles with rotation order X-Y-Z are calculated using the rotation matrix $\mathbf{R}_{IE}$ as stated in \cite{Siciliano2008}. ${_I}{\mathbf{v}}_{0E}$ and ${_I}{\mathbf{a}}_{0E}$ are the first and second derivatives of ${_I}\mathbf{r}_{0E}$, whereas ${_I}{\boldsymbol{\omega}}_{0E} = [\mathbf{e}_1 \ \ \mathbf{R}^T_{\alpha} \mathbf{e}_2 \ \ \mathbf{R}^T_{\alpha} \mathbf{R}^T_{\beta} \mathbf{e}_3] {_I}\dot{\boldsymbol{\varphi}}_{0E}$, using the elementary rotations $\mathbf{R}_{\alpha}$ and $\mathbf{R}_{\beta}$ \cite{Siciliano2008}, is the angular velocity and its derivative ${_I}\dot{{\boldsymbol{\omega}}}_{0E} = {_I}{\boldsymbol{\alpha}}_{0E}$ is the angular acceleration. The trajectories are plotted in \cref{fig:robotTrajSin2}. 
Additionally, the velocity ${_I}\mathbf{v}_{0T}$ from the point that is currently located at the TCP, which equals the velocity of the process orientation frame $\mathcal{F}_F$ along the processing path, is evaluated in \cref{fig:robotTrajSin2} to show that the desired processing speed is obtained in the respective directions. The mapping of each point on the processing path to one point on the final robot path may cause cusps dependent on the original geometry. To smooth the resulting path, it could be again evaluated as a new spline, which leads to reduced jerk, but increases the deviation in the area of the cusps. Another approach would be the determination of the path parameter using a constrained optimization problem considering jerk limits.

\section{Conclusion and Outlook}
Unlike the usual way of robot assisted processing of parts, where the robot holds the tool, in some cases it is easier to spatially fix the tool and handle the part instead. The presented approach showed a way to calculate a trajectory for the robot, where the desired operation point remains at a spatially fixed TCP, while maintaining a predefined process orientation. Therefore, only mathematical descriptions or some sampling points for the shape of the processing path on the part are given. The use of B-splines enables the path to be continuous with a desired degree. Suitable standard motion profiles for the arc length parameterized path parameter $\sigma$ allow to create a smooth robot trajectory considering maximum velocity, acceleration and jerk limits. Tests on the real system validated the functionality of the approach. Further research in this area would be the optimization of the path parameter considering joint and workspace limits instead of using a standard motion profile. Additionally an optimization of the robot mounting point on the part and the general positioning of the tool in the workspace can be done, which may reduce peaks in jerk or in the calculated path itself, respectively. Another aspect would be the improvement of the processing quality in a further step of the optimization based on information about the process itself. This could be for instance the consideration of a thermal model for automated tape laying processes, taking the heating power of the system and thermal properties of the used tape into account.

%
%
%
\bibliographystyle{splncs04}
\bibliography{RAAD2025_Rameder.bib}

\end{document}